\documentclass{bmvc2k}

\title{Explainable Deep Learning for Video Recognition Tasks: A Framework \& Recommendations}

\addauthor{Liam Hiley}{hileyl@cardiff.ac.uk}{1}
\addauthor{Alun Preece}{preecead@cardiff.ac.uk}{1}
\addauthor{Yulia Hicks}{hicksya@cardiff.ac.uk}{2}

\addinstitution{
 Crime and Security Research Institute\\
 Cardiff University\\
 Cardiff, UK
}
\addinstitution{
 School of Engineering\\
 Cardiff University\\
 Cardiff, UK
}

\runninghead{Hiley et al.}{Explainable Deep Learning for Video
Recognition Tasks}

\begin{document}

\maketitle

\begin{abstract}
The popularity of Deep Learning for real-world applications is ever-growing. With the introduction of high performance hardware, applications are no longer limited to image recognition. With the introduction of more complex problems comes more and more complex solutions, and the increasing need for explainable AI. Deep Neural Networks for Video tasks are amongst the most complex models, with at least twice the parameters of their Image counterparts. However, explanations for these models are often ill-adapted to the video domain. The current work in explainability for video models is still overshadowed by Image techniques, while Video Deep Learning itself is quickly gaining on methods for still images. This paper seeks to highlight the need for explainability methods designed with video deep learning models, and by association spatio-temporal input in mind, by first illustrating the cutting edge for video deep learning, and then noting the scarcity of research into explanations for these methods.    
\end{abstract}

\section{Deep Learning for Video}\label{video}
Deep Neural Networks (DNNs) have provided competent solvers for Image Recognition problems for a decade now. With early efforts in pattern recognition \cite{fukushima:neocognitron} paving the way for super-human performance in 2011 \cite{cirecsan:superhuman}. The recent surge in machine learning has brought with it the innovation to apply DNNs to complex problems in other domains such as bioinformatics, and particle physics \cite{lecun:deep_learning}. One such domain being spatio-temporal data, most commonly video data, in which the motion of objects is depicted in images over time via animation.
\\
\subsection{Datasets}
The main drive behind development in video and spatio-temporal deep learning has been human activity recognition. Many datasets have been compiled of human action in videos, either for classical classification/recognition tasks \cite{kay:kinetics,soomro:ucf101,kuehne:hmdb,caba:activitynet, abu:youtube8m}, sparse localisation/detection \cite{yang:trecvid, kuehne:breakfast, sultani:ucfcrime} or other. The collection of video data proves difficult, with most datasets up until recently being small \cite{soomro:ucf101, kuehne:hmdb}, 13000 and 7000 samples respectively, very short \cite{yang:trecvid, sultani:ucfcrime}, 150 and 128 hours respectively, or notably short \cite{kuehne:breakfast}, 77 hours total. What's more, while the number of samples would be considered small for an image dataset, the need for data is intuitively greater for video deep learning models, which consistently rely on larger sets of sometimes more complex parameters for inference.
\\
The development of the Kinetics dataset \cite{kay:kinetics} somewhat revived the field by providing a comparatively very large dataset, at 300,000 samples initially. Kinetics boasts a rich variety of human action, mined from youtube videos and labelled using their titles, that has proven useful for pretraining feature extractors for large models that results in a comprehensive feature space for application on smaller more specific tasks. Kinetics has since become an analogue to the benchmark ImageNet \cite{russakovsky-et-al:imagenet_2014}.
\\
Recently, similar efforts have been made to compile massive-scale video datasets of activity with \cite{abu:youtube8m} and \cite{monfort:moments}, although they have not yet found the same success as \cite{kay:kinetics}.

Other groups have made efforts solving the question of gathering large enough sets of data for video machine learning tasks. Namely by accessing video surveillance footage \cite{yang:trecvid, sultani:ucfcrime}, a rich and varied source of human activity that is for the most part organic and unrehearsed. This medium for human activity captures more of what's considered the mundane or ordinary activity that might be less common in a source such as YouTube. In \cite{sultani:ucfcrime}, researchers collected footage of crimes such as Abuse, Burglary and Assault that can be associated with one, continuous activity, with the hopes of providing data for anomaly detection, for which single-event crimes are perfect examples in a video stream mostly comprised of regular day-to-day activity.

\subsection{Convolutional Neural Networks, Recurrent Neural Networks and Long Short-Term Memory Models}\label{2dconvs}
Recently the jump has been made from Still Image Processing tasks to streams of images in video input, with a focus on Human Activity Recognition \cite{ji:3dconv,simonyan_zisserman:two-stream,karpathy:large,donahue:long_rnn,wang:temporal,wang:sr-cnn}. At first, attempts were made to pool together spatial features extracted from video frames, treated as individual images by applying successful image processing architectures, Convolutional Neural Networks (CNNs) \cite{karpathy:large}. The power behind CNNs lie in their kernel operators. The convolutional kernel acts as a sliding window, passing over an image and outputting the correlation between that section of the image and the kernels weights, which represent visual features such as lines and corners in the early layers, to scales on fish at much later layers. This technique is simple enough to adapt from image networks, and had the advantage of pretrained weights from massive scale image datasets such as \cite{russakovsky-et-al:imagenet_2014}. However, it fails to retain temporal information from the stream, essentially flattening the frame block into a set of unique instances.

To encode temporal information into the model's decision, each model in the frame-by-frame representation could be folded or spooled into a Recurrent Neural Network (RNN) \cite{donahue:long_rnn}, a method popular in natural-language processing and speech recognition that function by spooling together many networks, each for a point in time. By weighting the network outputs, the full model can effectively learn sequence data, and is trained by unfolding into a large DNN. LSTMs improve upon RNNs in that they can retain temporal information over longer sequences via their Long Short-Term Memory cells, which replace the original `cell' or network body in each time step's network. However, both are costly to train, with backpropagation flowing through the entire unspooled network. Both also require a large spatial feature extractor to translate the input frame to a feature representation.
\\
\subsection{3D Convolutions}
A video stream can be viewed as a ordered set of RGB frames, the temporal aspect of the video comes from the relationship between frames. By modelling the input data, not as a set of 2D data, but as a block of 3D data, temporal information is no longer a relation between inputs but is now inherent in the input. Each pixel becomes a voxel, encoding the colour intensity of a point in space and time. To match this, convolutional and pooling kernels need to be inflated to 3D \cite{ji:3dconv}. This method relies on a single-stream spatio-temporal feature extractor to provide a rich enough feature space to contain the extremely varied movements found in human activity (e.g. Skiing vs. Chopping onions).
\\
In \cite{ji:3dconv}, they present C3D, a small convolutional network with one fully connected layer for high-level reasoning. The input to the model is a 7 channel video, consisting of R, G, B frames as well as X and Y gradient and optical flow fields (discussed further in \ref{two-stream}). This 3D representation resulted in the network surpassing baseline methods, including \ref{2dconvs} on the TRECVID Human Action Surveillance dataset \cite{yang:trecvid}.
Since C3D, efforts have been made to retrace the success of popular CNN architectures from Image Recognition. In \cite{kensho-et-al:3dresnet}, members of the ResNet \cite{he-et-al:resnet} family of models were able to outperform C3D and other contemporary methods on the Kinetics \cite{kay:kinetics} and UCF-101 \cite{soomro:ucf101} action recognition datasets, showing that 3D Convolutions can be treated similarly to their 2D counterparts. The identifying feature of a ResNet model is it's use of residual skip connections, that connect input to a convolutional block to it's output so that the `signal' is strengthened in a manner.
\subsection{Two-Stream Networks}\label{two-stream}
Rather than learn sequences of features using a recurrent layer, Simonyan and Zisserman \cite{simonyan_zisserman:two-stream} implemented a two-stream approach that aggregates the decision from two models in order to classify the video.
\\
The first model, or stream is a regular CNN taking RGB images as input, similar to the early works using pooling from frame-by-frame image models. The second stream operates on the temporal dimension of the video, which is approximated via stacked optical flow frames. These optical flow fields can be taken to represent the motion between intermediate frames in a video, by encoding the apparent velocities between pixels using brightness as a measure of movement.
Like the RNN technique, this method has the benefit of considering the temporal aspect of the video, which intuitively must be considered for input that is temporal in nature. Two-Stream Networks are also in some respect recurrent, in that the optical flow is recurrently optimised. Two-stream approaches also have the added benefit over RNNs, of capturing finer low-level motion \cite{carreira:quo}. Two-stream architectures have since gone on to be improved in various ways \cite{wang:sr-cnn,wang:temporal} and optical flow is still considered beneficial to activity recognition problems to date, with two-stream approaches even taken to 3D CNNs \cite{carreira:quo}. This two-stream, 3D CNN, dubbed I3D, named for its pretraining method, inflating ImageNet weights from 2D, averages Stacked Optical Flow and RGB trained models at test time. This achieves state-of-the-art performance on \cite{kay:kinetics}, \cite{soomro:ucf101} and \cite{kuehne:hmdb}. In \cite{wang:hallucinating}, it is shown to be possible to `hallucinate' optical flow fields for still image input using a Hybrid Video Memory machine. The HVM uses a memory of similar still images to the input to generat an optical flow for the input from noise, based on the relevant optical flow fields given to it with the memorised images. Therefore, for a still image of an action/activity, the model could relatively cheaply and successfully generate temporal features that would then be used as input for the second stream of a two-stream architecture, whereas before the decision would rest solely on the RGB channels.

\section{Explaining spatio-temporal models }\label{explaining}

\subsection{Kernel visualisation and Global Explanations}
Efforts to explain deep learning models developed for video tasks often seem to be included as an afterthought, an additional justification of the method to supplement the benchmark performance. 
One logical method is to visualise the kernels, that learn the spatio-temporal features that make these models unique. Kernel visualisation is common in Image tasks as a method of linking deep learning decisions. In \cite{erhan:visualizing}, images were optimised to maximally activate single filters at a time in one of two deep learning models trained on the MNIST handwritten digit dataset \cite{lecun:mnist}. This was adapted to Convolutional networks in \cite{simonyan:conv_visualising}. \cite{olah-et-al:building} provides an interactive and comprehensive investigation of feature visualisation for a large deep Image recognition model.
In \cite{simonyan_zisserman:two-stream}, the previous method was adapted for filters trained on optical flow, in an effort to justify the model by way of linking the features it learns to previous hand-crafted methods. In \cite{carreira:quo} the method is used to visualise the 3D filters, slice by slice rather than as cubes, with the aim of showing the richness of the filters. Again, the aim of these works is not to investigate the features learned by the networks, but to propose the networks themselves.
In \cite{wang:temporal}, Deep Draw is applied to a Temporal Segment Network, a Two-Stream Network that adapts sparse sampling to classify using the entire video. The result is an image optimised from noise that maximally activates a particular neuron in the logits, or final, layer, relating to an output class. The authors of DeepDraw \cite{audun:deepdraw} note the issues of optimising on noise without constraint, mainly that it generates crowded images that are difficult to decipher.
The use of DeepDraw in \cite{wang:temporal} gives insight into the models understanding of the classes identity in the feature space. Showing intuitive associations between motion and activities, such as ripples in water with the activity of Diving, in the optical flow stream which emphasises the influence of moving bodies of water on the decision. The spatial stream for the same class however activates much more strongly it seems on the presence of humans, topless or in swimsuits. 
Interestingly, \cite{tran:spatiotemporal}, a work on spatio-temporal feature learning with the C3D \cite{ji:3dconv}, did not use this method, preferring instead Deconvolution \cite{zeiler:visualizing} which visualises a filters activations on a given input sample. This local explanation of the learned features, in the \cite{lipton:mythos,ribeiro:lime} sense of the term local, is used to infer how such features have developed. They note that the C3D filters at first focus on the spatial dimension but quickly transfer focus to the motion in the frame, the temporal dimension. Also notable, is that \cite{hara:spatiotemporal}, somewhat of a successor of \cite{tran:spatiotemporal}, does not attempt to visualise the features it sets out to learn. An application of feature visualisation, in the spirit of explainability can be found in the work of Nerinovsky \cite{arseny:3d}. Here the kernels at different layers of the popular I3D network \cite{carreira:quo} are visualised, producing animated videos of swirling features coming in and out of focus. Unlike the recognisable eyes, noses and other such features of animals, it is hard to gain an intuition from these sliced cubes as to what each might detect in a video of human activity, although at earlier layers it is clearer to determine the general type of motion that activates each kernel.

While still explaining the model using it's weights, \cite{kim:tcav} takes a very different approach to visualisation. Concept Activation Vectors (CAV) are partially handcrafted features for model explanations. CAVs are generated by manually segmenting the model's training dataset into two groups, such that one group contains all samples that represent or at least contain instances of a `concept' as defined by a human. A linear model is then trained on the two subsets, where the input is the activations of the weights of the model on the samples from each subset. The concept activation vector is then the normal of the hyperplane separating the two subsets, towards the concept class. This results in an explainer model that can flag the model as having seen a concept in it's input, based on the models activations for that input. The authors have so far only applied this method to image data, but suggest video as one of a few possible extensions.

\subsection{Explaining Local Decisions in the Input Space}
Local explanations, explanations that explain the decision on a single sample, have found significant success in the Image domain, with most of the techniques developed with that input medium in mind. Sensitivity analysis \cite{baehrens:sensitivity, rasmussen:sensitivity} can be considered one of the earliest attempts at local, visual explanation with gradients of class probability, vectors in the direction of the decision boundary, are displayed on digit classification to illustrate how to change the digit 2, for example, so that the model thinks it's an 8. While useful and intuitive for simple examples like digit classification, sensitivity analysis often gets noisy for larger more complex problems.
\\
\subsubsection{Layer-wise Relevance Propagation and White-box methods}
Increasingly popular are the Layer-wise Relevance Propagation (LRP) family of techniques, first defined in \cite{bach:lrp} which take a white-box approach to local explanations, in that they assume access to the model internals . LRP is proposed as a method for pixel-wise decomposition of relevance to a decision. Although multiple variations on the method are presented, the main theory centres around the following conditions that must be satisfied for the explanation to be considered an LRP explanation.
\begin{itemize}
    \item{The relevance at each layer must sum to the output of the model.}
    \item{The relevance at any neuron in the layer other than the output layer is the sum of incoming relevances to that layer.}
\end{itemize} 
The authors clarify that this can be satisfied by meaningless explanations and as such the formula for LRP should be taken more as a framework to follow in developing explanation methods. The work has since been adapted and implemented by other researchers \cite{shrikumar:deeplift, zhang:eb}, and maintained and improved by the original authors \cite{binder:lrp}. In \cite{montavon:deeptaylor} the original authors propose an implementation of their prior theoretical work based on Taylor expansions of the decision function at a pre-defined root point. The relevance of a neuron is then defined locally in comparison to the root point. This method has since been applied in various fields for explaining deep models \cite{schutt:schnet, schirrmeister:eeg, arras:text, srinivasan:videolrp}. Note: \cite{srinivasan:videolrp} is an application of the LRP method to compressed domain human action recognition. And as such while not a method covered in section \ref{video}, is still a very rare example of an investigation of explainability in the spatio-temporal domain.

In \cite{sundararajan:axiomatic}, the authors note that two models that provide exactly the same output for all inputs, regardless of model internals or implementation, should provide identical explanations, an axiom known as Implementation Invariance. By implementing discrete intermediate gradients, for which the chain rule cannot be applied, \cite{bach:lrp} and \cite{shrikumar:deeplift} are not effective explanation methods, as relevance in the implementation may be attributed even though it is not relevant in the input. The authors go on to define integrated gradients, a method that instead integrates the gradient of the model function against a root point, at all intermediate points. This method is then shown to satisfy the completeness rule, that the relevance sums to the difference between the input and the root, that is defined as desirable in \cite{bach:lrp}.

LRP methods are inherently applicable to the spatio-temporal domain, since they attribute relevance from the output onto the input space, irrespective of dimensionality. Describing the relevance of the temporal dimension on the decision at a voxel is however still only implied through difference in similar spatial features between frames. That is, by reconstructing the relevance field into a video, it is difficult to see whether an area in a frame is attributed high relevance because of its spatial information, or because of its temporal information. In \cite{srinivasan:videolrp} they instead opt to show the amount of relevance over time, including the frames as reference at key points. This better shows the effect of motion on the decision, which suggests that explanations for video, cannot be digested in the same way that the inputs are.

Other white-box methods have been developed that do not use the LRP framework. CAM \cite{zhou:cam} notes the use of Global Average Pooling, originally developed for training regularisation, for localisation of feature map activations. This produces a heatmap of the areas in the image that activated strongly to a particular class. The authors note the general loss of spatial information in hidden layers, and thus this technique is not compatible with models that make use of such layers. Grad-CAM \cite{selvaraju:gradcam} addresses this by first propagating the gradient of the class output node back through the hidden layers, with respect to the last convolutional layers activations. From then on CAM can be applied. CAM and Grad-CAM produce class discriminative heatmaps that perform very well on object localisation without any additional training. This has the added benefit of showing objects in the scene that the model recognises as relevant to the decision. These, again, are applicable to video data, much moreso in the case of \cite{selvaraju:gradcam} since the majority of models for video tasks use at least one fully connected layer.

The SHAP (SHapley Additive exPlanations) framework for local explanations \cite{lundberg:shap} finds similarities in \cite{shrikumar:deeplift, ribeiro:lime, bach:lrp} and proposes a framework based on methods from game theory, to improve all methods. SHAP explanations are based on multiple explainers for different models. The deep explainers are gradient based and white-box. The authors suggest that SHAP values are the only possible consistent, and locally accurate additive feature attribution method.

Perhaps the least efficient but most model-faithful explanation technique is proposed in \cite{ross:right}, where models are trained with a term for explainability included in the loss. The authors constrain input gradients on features considered to be irrelevant to the decision. This discourages the model from learning on those features, which should result in a model that focuses more on objects in the scene that a human deems significant.

To our knowledge, the only explanation method in this format developed solely for video can be found in \cite{stergiou:tubes}, and \cite{hiley:discdtd}. In \cite{stergiou:tubes} the authors map the activations to the input in a way that is similar theoretically and visually to \cite{selvaraju:gradcam}. This method, known as Saliency Tubes, provides cylindrical heatmaps that visualise the focus of attention in the input video, through each frame. As an alternate visualisation, the stack of frames is staggered so that the path of the tube through the video can be seen at once. This method for visualising the motion is useful in better translating the temporal aspect of the explanation without animating the frames. In \cite{hiley:discdtd}, the authors suggest separating the relevance generated by the deep Taylor method, for 3D models, into its spatial and temporal components. They also provide a novel yet simple way of approximating this that shows that models do seem to attribute relevance to some objects in a scene because of their motion, much more so than their appearance.

\subsubsection{Black-box methods}
Other efforts have been made to develop explanations for models without access to the model internals. The most popular of these techniques, known as Local Interpretable Model Explanations or LIME \cite{ribeiro:lime}, seeks to approximate the decision function by many closely sampled input points, which all center around the input point to be explained. It can then attribute positive or negative influence on the decision function to the differences in the sampled inputs, and overlay this on the original input. LIME has found much success and support in the field, and is implemented for text, image, and tabular data. In \cite{stiffler:lime++} they note the instability of LIME: Since it uses random seeding to draw samples, via segmentation in images, multiple explanations of the same input for the same model can result in very different attributions. The authors suggest aggregating explanations, to improve stability. 

The theory behind LIME itself is extendable to all popular machine learning input domains. The issue lies in how the samples are drawn from the input. For text processing, words can be considered atoms to be used for sampling, the same can be said for items in tabular data. LIME's solution to sampling images is based on segmentation. This again results in instability, and would for any 3D segmentation adapted for video inputs.

\subsubsection{Explanation by example}
In \cite{koh:influence}, a different approach to local explanations is taken. Instead of attributing relevance to features in the input sample, the implementation of which all previous methods \cite{bach:lrp,shrikumar:deeplift, montavon:deeptaylor, ribeiro:lime, lundberg:shap} have argued over, Influence Functions instead explains the model's decision on an input in terms of the training samples that most influenced that decision. They do this by discretely deriving the loss function at the explained sample, with respect to each input in the training data. This is computationally massive task to achieve, especially in the chosen example domain of images, with large datasets like ImageNet. However, the authors also show that even approximations to the derivative can provide insight into what influenced the decision.

This method does not attempt to alter the input, and only relies on being able to provide that input as an example. Therefore, it is the most naturally applicable to the video domain. The explanation can be presented in the same manner as the model input without any processing involved.

\section{The Interpretability of Explanations}
While providing an explanation that is faithful to the model, i.e. accurately represents the models entire decision process, is a quality sought after by many in the literature \cite{lundberg:shap, montavon:deeptaylor, audun:deepdraw, erhan:visualizing, zeiler:visualizing} (\cite{ross:right} by far the most so), the resulting explanations are often noisy and sometimes indecipherable to a human, showing that transparency is not necessarily the only quality to optimise for when generating explanations. Other methods seek instead to be more digestible for humans \cite{ribeiro:lime, kim:tcav}. Many works have sought to provide guidelines on more interpretable explanations, in recent years. In \cite{ribeiro:lime}, they note the tradeoff between an explanation's fidelity to a model and its interpretability to humans. Lipton \cite{lipton:mythos} defines some qualities of an Interpretable model, namely transparency, and post-hoc interpretability, which covers the explanation methods noted above (Section \ref{explaining}). He goes on to note a few key cautions to take when optimising interpretability: Linear models are not necessarily more interpretable than deep networks, Transparency can contend with performance if parameters are limited so that they are more understandable. Also, that a plausible explanation, i.e. one that perfectly highlights a salient object, might not be a true explanation of the model, and so should not be the only factor when grading that explanation. It is suggested in \cite{dhurandhar:tip} that interpretability is not necessarily only human understanding. This work also introduces $\delta$-interpretability, or the quality of a model being able to improve performance on a task by way of explanation.
Interpretability and Explainability are first disinguished in \cite{tomsett:interpretable}, where Explainability is the capability of the model to provide an explanation that is faithful to the causal factors of that decision. Interpretability is defined here as the information on the decision an agent can gain by use of the explanation, given the transparency of the model. This sets up Interpretability as a qualitative measurement for Explanations. The authors go on to suggest that since the role an agent plays in the ecosystem formed around a supporting AI, Interpretability is subjective to the agent. Context such as field knowledge, motivation and time constraints must factor into the explanation provided, as different information is valuable based on the use the agent intends to find from the explanation. 
There is still disagreement on how to quantify the explainability and interpretability of an explanation, \cite{hoffman:metrics} suggests guidelines for measuring explainability, that would also assure interpretability in the above sense. In \cite{samek:metric} attribution methods are quantitavely assessed by ordering the pixels of the relevance map and then repeatedly removing the first most relevant pixel. By degrading the image in this way, the authors theorise that explanations that best represent the decision will cause the greatest drop in performance for that sample when passed again through the model. This method confirms that \cite{bach:lrp} outperforms \cite{rasmussen:sensitivity} and \cite{zeiler:visualizing} in model fidelity. This method's accuracy is called into question in \cite{hooker:roar}, where it is suggested that removing information from an image, results in artifacts that essentially shifts the distribution of input features to one that the model was not originally trained on. Therefore, it is suggested instead to retrain the model after removal of the feature, and test the models performance then. It was shown through this that models are surprisingly robust to feature removal, and can still reliably classify images with very little information left from the original sample.

\section{Discussion}
It seems the main issues with interpretability and explainability of Deep Learning models for video tasks is first and foremost a lack of attention. The scope of explainable AI is still very much centered on the Image domain, and efforts in explaining spatio-temporal models do not attempt to adapt techniques to the new modality. Work to highlight the hidden temporal nature of explanations past animating the frames, in \cite{srinivasan:videolrp, stergiou:tubes} currently suggests unraveling the video into its frames and their explanations, but this is then either cluttered (as in \cite{stergiou:tubes}) or loses the videos sense of sequence. Future work on explaining motion in video will likely use separate explanations of temporal saliency on optical flow fields, as these networks are still widely used. However, developing a system in which spatial and temporal relevance do not interfere with one another when displayed together in one image, would provide a visually efficient mode of explanation that does not require referencing between multiple images, and would allow the explanation to be viewed in the same way as its input. Explanation could also add insight into the influence of motion on model decisions. The main approaches to video tasks are thought to capture temporal information at different abstractions \cite{carreira:quo}, explanations that are heavily motion focused, with little relevance attributed to spatial features, would support this statement.
\section{Framework for Future Research}
In this paper we have addressed some gaps in the current works for explainable activity recognition with video deep learning models. To suggest some possible solutions for future efforts, we propose a framework for improving this field, from it's current state:
\begin{enumerate}
    \item Many of the current approaches to explainability for video models are adapted from methods for still image models. To faithfully explain video models, it will be necessary to design new methods intended primarily for these models.
    \item Applications of video models in a deployment setting will often feature real-time processing, for example in a surveillance scenario. This is necessary to take into account when implementing explanation methods, since heavier-weight methods would be infeasible in such a scenario.
    \item Furthermore, considering applications must be taken into account, to ensure the success of the method. Explaining models with complex representations of motion such as 3D CNNs and C-RNNs can result in hard to interpret results. Explanations that aim for transparency and faithfulness might fall short for a user with no background in machine learning or computer vision. This problem is addressed in \cite{tomsett:interpretable}, where the authors design a framework for the user roles in a explainable deep learning system. They highlight the need to consider a user's background and their motivation for requesting the explanation, when providing that explanation. This should be the case for future video deep learning explanation methods as well.
\end{enumerate}
\section{Conclusion}
In summary, video deep learning is currently at the forefront of machine vision, with a rich and varied body of work committed to comprehensively and compactly representing time in imagery. However this success and interest has not attracted much work in explaining these models, in contrast to image recognition for which Explainable AI is likely now most popular. Visualising the features learned by these spatio-temporal models has been thought of by some when designing new architectures, but mostly to justify that the models are learning \emph{something}, which is an observation oriented towards a researcher and perhaps less useful when justifying a model for deployment. Local explanation methods have also been extended to these models, but it is clear that out of the box these methods are very much anchored to their origins in the image domain and as such seem an ill fit to the video input. Current work in adapting popular explanation methods such as visualisation \cite{arseny:3d}, deep Taylor \cite{srinivasan:videolrp} and Grad-CAM \cite{stergiou:tubes} show promise and possibility for understanding these models, and there are clear directions to take in the future such as exploring the use of optical flow, in the explanation as much as it is already in the decision, as well as a method to highlight temporally-salient regions in an explanation in order to distinguish the models decision making process from that of a similar model learned on images. 

We provide a framework for addressing these problems, notably the necessity to develop native video explanation methods, that these methods should be lightweight enough to run near real-time, and that the developers take into consideration the intended users and their usecases for these techniques.
\newpage


\end{document}